\documentclass{article}
\usepackage{PRIMEarxiv}
\usepackage[utf8]{inputenc}
\usepackage[T1]{fontenc}
\usepackage{hyperref}
\usepackage{url}
\usepackage{booktabs}
\usepackage{amsfonts}
\usepackage{nicefrac}
\usepackage{microtype}
\usepackage{algorithm}
\usepackage{algpseudocode} 
\usepackage{fancyhdr}
\usepackage{graphicx}
\graphicspath{{media/}}
\usepackage{amsmath}

\makeatletter
\def\@fnsymbol#1{}
\makeatother

\title{A Blueprint for Self-Evolving Coding Agents in Vehicle Aerodynamic Drag Prediction
\thanks{\textsuperscript{1}{Famou Agent Team, Baidu AI Cloud} \quad \textsuperscript{2}IAT AI Team}}

\author{
  \textbf{Jinhui Ren$^{1}$,} \enspace
  \textbf{Huaiming Li$^{1}$,} \enspace
  \textbf{Yabin Liu$^{2}$,} \enspace
  \textbf{Tao Li$^{2}$,} \enspace
  \textbf{Zhaokun Liu$^{2}$,} \enspace
  \textbf{Yujia Liang$^{2}$,} \\
  \textbf{Zengle Ge$^{1}$,} \enspace
  \textbf{Chufan Wu$^{1}$,} \enspace
  \textbf{Xiaomin Yuan$^{1}$,} \enspace
  \textbf{Danyu Liu$^{1}$,} \enspace
  \textbf{Annan Li$^{1}$,} \enspace
  \textbf{Jianmin Wu$^{1}$} \\
  \vspace{0.2cm} \\
  \textsuperscript{1}\textbf{Famou Agent Team, Baidu AI Cloud} \quad \textsuperscript{2}\textbf{IAT AI Team} \\
}

\begin{document}
\maketitle

\begin{abstract}
High-fidelity vehicle drag evaluation is constrained less by solver runtime than by workflow friction: geometry cleanup, meshing retries, queue contention, and reproducibility failures across teams. We present a contract-centric blueprint for self-evolving coding agents that discover executable surrogate pipelines for predicting drag coefficient $C_d$ under industrial constraints. The method formulates surrogate discovery as constrained optimization over programs, not static model instances, and combines Famou-Agent-style evaluator feedback with population-based island evolution, structured mutations (data, model, loss, and split policies), and multi-objective selection balancing ranking quality, stability, and cost. A hard evaluation contract enforces leakage prevention, deterministic replay, multi-seed robustness, and resource budgets before any candidate is admitted. Across eight anonymized evolutionary operators, the best system reaches a Combined Score of 0.9335 with sign-accuracy 0.9180, while trajectory and ablation analyses show that adaptive sampling and island migration are primary drivers of convergence quality. The deployment model is explicitly ``screen-and-escalate'': surrogates provide high-throughput ranking for design exploration, but low-confidence or out-of-distribution cases are automatically escalated to high-fidelity CFD. The resulting contribution is an auditable, reusable workflow for accelerating aerodynamic design iteration while preserving decision-grade reliability, governance traceability, and safety boundaries.
\end{abstract}

% keywords can be removed

\section{Introduction}
Vehicle aerodynamic drag prediction remains a classical engineering task with a distinctly modern bottleneck: \emph{iteration throughput}. In production programs, teams must screen hundreds to thousands of geometry variants under coupled constraints (thermal management, packaging, manufacturability, regulations, and schedule). In that regime, the limiting factor is rarely solver access alone; it is the end-to-end latency from geometry generation to decision-ready evidence. This pressure is even stronger for electric vehicles, where small improvements in $C_d$ can materially affect range and energy efficiency \cite{taira2024ev}.

High-fidelity CFD remains the judge of record, but the true cost of one usable result extends far beyond solver wall time. Engineers must handle CAD cleanup, meshing retries, setup tuning, convergence supervision, reruns, and post-processing across heterogeneous toolchains \cite{witherden2024pyfr,rumsey2025}. In practice, pipeline fragility---especially in meshing and data handoff---often dominates total effort, so program velocity is constrained by workflow reliability rather than raw compute capacity \cite{meshgenaisurvey2025,automeshgen2025}.

This reality reframes the objective from point prediction to \emph{decision support}. The central question is not only ``can we estimate $C_d$?'' but ``can we rank candidate designs reliably enough to allocate limited CFD budget to the right subset?'' Ranking errors directly translate into wasted queue slots, delayed milestones, and reduced trust in surrogate-assisted development.

Existing surrogate research spans classical statistical modeling, multi-fidelity learning, and modern geometry-aware deep architectures. However, deployment gaps persist in aspects that benchmark-centric reporting often underemphasizes: split integrity, leakage prevention, reproducible execution, and conservative behavior under distribution shift.

We therefore argue that practical deployment requires two system-level commitments. First, a strict \emph{evaluation contract} must govern the full executable pipeline---including split policy, multi-seed robustness, resource limits, and artifact traceability \cite{smrs2025,atlasml2025,reprodl2025}. Second, operation must be uncertainty-aware: low-confidence or out-of-distribution inputs are escalated to CFD instead of being forced through overconfident extrapolation \cite{chu2023uqcfd,uq4cfd2024,surrelem2019,practicalguideuq2025,rezaeiravesh2021uqit}.

Against this background, we present a blueprint for a \emph{self-evolving coding agent} that searches over versioned, executable surrogate programs rather than tuning a single static model. Building on agentic tool-use paradigms \cite{yao2022react,schick2023toolformer,yang2024sweagent,feurer2020autosklearn2,fang2025survey} and extending Famou-Agent-style evaluator-guided evolution \cite{li2025fmagent}, the framework combines population-based exploration, contract-aware selection, and failure-driven refinement in one auditable loop.

\textbf{Contributions.}
\begin{itemize}
  \item \textbf{Industrial framing with explicit safety posture:} we cast drag prediction as a workflow-level engineering problem with concrete failure surfaces (export/mesh/solve/QC/post-process), and specify where surrogate outputs are acceptable versus where CFD escalation is mandatory \cite{witherden2024pyfr,rumsey2025,meshgenaisurvey2025}.
  \item \textbf{Contract-centric Famou-Agent workflow:} we organize Famou-Agent-style tool use, execution, memory, and evaluation under a fixed contract so progress is measured by reproducible criteria rather than anecdotal gains \cite{li2025fmagent,smrs2025,atlasml2025}.
  \item \textbf{Evolutionary coding-agent methodology:} we formulate surrogate discovery as population evolution over executable pipelines (islands, migration, and reliability-aware fitness), enabling structural innovation under industrial constraints.
  \item \textbf{Reusable deployment blueprint:} we provide an audit-oriented evaluation template and reporting protocol for vehicle drag surrogates, suitable for both public benchmarks and proprietary engineering datasets.
\end{itemize}

\textbf{Paper roadmap.} Section~\ref{sec:related} reviews prior work across four axes: aerodynamic surrogates, agentic AutoML, evolutionary program synthesis, and reproducibility governance. Section~\ref{sec:industrial} details the industrial workflow and risk model. Section~\ref{sec:fmagent} introduces Famou-Agent and the evolutionary coding-agent perspective. Section~\ref{sec:application} specifies the surrogate application, data protocol, and evaluation contract. Section~\ref{sec:analysis} reports performance and attribution, followed by deployment impact with integrated safety boundaries (Section~\ref{sec:impact}).

\section{Related Work}\label{sec:related}

Recent progress relevant to this paper spans four intertwined streams: aerodynamic surrogate modeling, agentic AutoML, evolutionary program synthesis, and deployment governance. We review each stream with emphasis on the remaining integration gap in industrial vehicle design.

\subsection{Aerodynamic Surrogates: From Statistical Models to Geometry-Aware Learning}
Classical surrogates (e.g., Kriging, RBF, polynomial response surfaces, and Gaussian-process variants) remain effective when design spaces are low-dimensional and sampling budgets are limited \cite{chen2018kriging,surrelem2019,gp2024}. Multi-fidelity formulations and MLMC-style estimators improve sample efficiency by combining abundant low-fidelity data with sparse high-fidelity labels \cite{multifidelity2025,pisaroni2017mlmc}. These approaches are still attractive for early design-space screening, but they scale poorly as geometric complexity and variable count increase.

Deep learning surrogates address that scalability challenge by operating on richer geometry representations. The field has moved from 2D image encodings \cite{song2023surrogate} to point-cloud and mesh/graph architectures \cite{singh2026car,nabian2024xmeshgraphnet,pfaff2021meshgraphnets}. This transition is enabled by larger public benchmarks. Beyond scalar $C_d$ prediction, recent models also infer richer flow quantities \cite{bleeker2025neuralcfd,hagnberger2026smart}.

Neural operators further extend this line by learning mappings between function spaces rather than fixed discretizations. FNO- and DeepONet-style methods can provide resolution-transferable surrogates for PDE-governed systems \cite{li2021fno,lu2021deeponet,pathak2022fourcastnet,takamoto2022pdebench,shukla2024deep}. In principle, this makes them well suited for aerodynamic workflows with changing meshes, although practical adoption still requires robust automation and careful deployment controls.

\subsection{Coding Agents and Agentic AutoML for Scientific ML}
AutoML has evolved from algorithm/hyperparameter selection toward full-pipeline automation. Early systems such as Auto-Sklearn 2.0 and modern AutoML taxonomies established strong baselines for structured search and ensembling \cite{feurer2020autosklearn2,hutter2024automl,shen2024automl}. With LLMs, the frontier shifts to coding agents that can reason, call tools, write code, and iteratively repair workflows.

General paradigms such as ReAct and Toolformer \cite{yao2022react,schick2023toolformer}, together with engineering-focused agents like SWE-agent and OpenHands \cite{yang2024sweagent,wang2024openhands}, show that language agents can execute multi-step software tasks. Domain-specific systems, including Famou-Agent and emerging Agentic-SciML frameworks, indicate that these capabilities transfer to scientific pipelines when execution and evaluation loops are explicitly structured \cite{li2025fmagent,foameagent2025,scimlagents2025,agenticsciml2025,codexscientist2025,robeyns2025selfimproving,fang2025survey}. For aerodynamics, this motivates shifting from one-off model tuning to continuously improvable executable workflows.

\subsection{Evolutionary Program Synthesis and Quality-Diversity Search}
Evolutionary program synthesis provides a natural mechanism for searching over executable pipeline variants rather than static parameter vectors. Recent LLM-guided methods such as FunSearch demonstrate how code proposals can be generated and selected in iterative loops \cite{romeraparedes2024funsearch}. Quality-diversity methods (e.g., MAP-Elites) preserve behavioral diversity while pursuing high fitness, helping avoid premature convergence in large design spaces \cite{mouret2015mapelites}. Multi-objective optimizers such as NSGA-II remain useful when accuracy, stability, and computational cost must be traded simultaneously \cite{deb2002nsgaii}. Large-scale code-generation systems (e.g., AlphaCode) further support the feasibility of search-driven program discovery \cite{li2022alphacode}.

For industrial surrogate discovery, these ideas suggest an ``evolve-and-select'' paradigm in which populations of candidate training/evaluation programs are continuously mutated, stress-tested, and pruned. This is especially relevant when requirements include not only predictive quality but also deterministic replay, bounded runtime, and integration readiness.

\subsection{Reproducibility, Governance, and Trust in Industrial Deployment}
Deployment in engineering organizations requires stronger guarantees than benchmark-level accuracy. Reporting and governance frameworks such as SMRS, Atlas, and RepDL emphasize dataset versioning, execution traceability, and reproducible experimental records \cite{smrs2025,atlasml2025,reprodl2025}. Datasheets and model cards complement these standards by documenting intended scope, limitations, and risk assumptions \cite{gebru2021datasheets,mitchell2019modelcards}. In the automotive domain, SAE-oriented guidance further stresses uncertainty quantification, validation discipline, and fail-safe operational policies \cite{sae2024ml,wong2025inductive}.

Taken together, prior work demonstrates strong ingredients but limited end-to-end integration. Existing approaches typically optimize static metrics (e.g., MAE) on fixed splits, whereas production drag-screening requires ranking reliability, uncertainty-aware escalation, and contract-level reproducibility. This paper targets that gap by combining evolutionary coding agents with an explicit industrial evaluation contract.

\section{Industrial Motivation: Drag Prediction in the Loop}
\label{sec:industrial}

\subsection{The Hidden Friction of the Simulation Loop}
In the idealized view of computational fluid dynamics (CFD), an engineer provides a geometry, a solver computes the flow field, and a drag coefficient emerges. In industrial practice, however, the cost of ``one CFD result'' is rarely measured in solver wall-clock time alone. Instead, it is a composite of manual labor, computational overhead, and organizational coordination \cite{witherden2024pyfr,rumsey2025}. For a typical automotive program, a high-fidelity RANS or DDES run may involve meshes on the order of $10^7$ to $10^8$ cells, requiring significant HPC resources and specialized expertise to manage.

The process typically begins with CAD cleanup, a notoriously labor-intensive stage where geometric artifacts---sliver surfaces, non-manifold edges, or misaligned tolerances---must be manually repaired to ensure meshability. Even with modern tools, meshing remains a primary bottleneck; a single complex geometry may require multiple retries as the mesh generator fails to resolve thin gaps or high-curvature regions, leading to a ``meshing tax'' that can consume days of engineering time before a single cell is solved \cite{meshgenaisurvey2025,automeshgen2025}. This tax is compounded by the need for local refinement zones around mirrors, wheels, and underbody components where flow separation is critical.

Once a mesh is successfully generated, the simulation enters the HPC queue. Here, the friction is not just FLOPs but contention. Engineers often face queue times of 6--36 hours for high-fidelity runs, creating a disjointed workflow where the feedback loop for a design change is measured in days rather than minutes. During this wait, ``solver setup drift'' can occur---minor variations in boundary conditions, turbulence model constants, or solver versions across a team can lead to inconsistent results that necessitate expensive re-runs. The rare-but-expensive failure---a case that diverges after 48 hours of compute or a boundary condition mismatch discovered only during post-processing---often dominates the critical path of a vehicle program.

\subsection{The Five-Stage Pipeline and Its Failure Modes}
To understand the fragility of the industrial loop, we must decompose it into its constituent stages. Each stage represents a potential point of failure that can reset the engineering clock. In our experience, the transition between CAD and mesh is the most frequent source of delay, but the late-stage failures in convergence or post-processing are the most costly in terms of wasted FLOPs.

\begin{itemize}
    \item \textbf{CAD Export \& Cleanup:} Failure modes include geometric artifacts and non-manifold edges. Typical time cost: 2--8 hours. Recovery: Manual geometry repair.
    \item \textbf{Meshing:} Failure modes include cell skewness, negative volumes, or failure to resolve gaps. Typical time cost: 4--24 hours. Recovery: Local refinement or CAD simplification.
    \item \textbf{Solver Setup:} Failure modes include boundary condition mismatch or incorrect turbulence model selection. Typical time cost: 1--4 hours. Recovery: Configuration audit and re-submission.
    \item \textbf{QC \& Convergence:} Failure modes include numerical divergence or oscillatory residuals. Typical time cost: 12--48 hours. Recovery: Step-size reduction or mesh coarsening.
    \item \textbf{Post-processing:} Failure modes include script errors or inconsistent integration surfaces. Typical time cost: 1--2 hours. Recovery: Re-integration of surface pressures.
\end{itemize}

This cascade implies that any surrogate model intended for industrial use must not only be fast but must also bypass as many of these fragile stages as possible. A model that requires a high-quality mesh to produce a prediction still carries the ``meshing tax'' of the original pipeline, limiting its utility in the early, high-velocity stages of design.

\subsection{The Coordination Tax and Version Drift}
Beyond the physics, there is a hidden coordination tax. Large-scale industrial projects involve multiple teams (aerodynamics, structures, thermal) often working on slightly different versions of the same master geometry. Ensuring that the drag prediction used for range estimation matches the latest structural reinforcement is a constant battle against version drift. Post-processing scripts, too, suffer from versioning issues; a change in the integration method for surface pressure can shift drag counts by a margin that masks real aerodynamic improvements \cite{sae2024ml,taira2024ev}.

In a production environment, the ``digital thread'' is often frayed. A change in a wheel arch liner by the packaging team might not be communicated to the aerodynamics team until after a 48-hour simulation has already been queued. This lack of synchronization leads to ``ghost'' drag changes where improvements in one area are offset by undocumented regressions in another. The implication for machine learning (ML) adoption is straightforward: a surrogate model is only valuable if it reduces the \emph{total} loop time---from CAD to decision---without introducing hidden risk. Replacing a 10-hour solver run with a 10-second ML inference is transformative, but only if the ML model does not require the same manual CAD cleanup and meshing steps that it was intended to bypass.

\subsection{Why Optimization Is Comparison-Heavy: Ranking vs. Absolute Error}
Aerodynamic optimization is fundamentally comparison-heavy. Engineers build evidence across dozens to thousands of controlled comparisons---sweeping through winglet angles, spoiler heights, or underbody curvatures---and rarely rely on a single decisive experiment. In this regime, the absolute value of the drag coefficient ($C_{\mathrm{d}}$) is often less important than the \emph{delta} between designs.

This shifts the fundamental requirement of the model: we need a predictor that is accurate enough to rank designs and quantify relative gains, and stable enough to avoid chasing numerical noise. A model with a low Mean Absolute Error ($\mathrm{MAE}$) but unstable ordering---where it incorrectly predicts that Design A is better than Design B due to sensitivity to mesh resolution or random seeds---is operationally dangerous. In a production environment, ``ranking reliability'' beats raw $\mathrm{MAE}$ in decision value. If a surrogate incorrectly identifies a sub-optimal design as the winner, the subsequent high-fidelity validation (which is always performed for the final selection) will reveal the error, but the engineering time spent exploring that dead-end is unrecoverable.

From a decision theory perspective, the value of a surrogate is tied to the expected value of information ($\mathrm{EVI}$). A surrogate that provides a noisy but fast estimate allows for a broader exploration of the design space, effectively increasing the probability of finding a global optimum. However, the cost of a ``false positive''---a design that appears promising in the surrogate but fails in CFD---must be weighed against the speed of iteration. In our experience, industrial reports suggest that a ranking correlation (e.g., Spearman's $\rho$) above 0.9 is typically required for a surrogate to be trusted in a production optimization loop \cite{surrelem2019,ashton2024windsorml}.

\subsection{When Is a Surrogate Safe to Deploy?}
The deployment of ML surrogates in the industrial loop is not an all-or-nothing proposition. Instead, it follows a risk-based hierarchy. For parametric tweaks, such as adjusting a spoiler angle by a few degrees, the design space is well-behaved and the risk of topological surprise is low. In these cases, a surrogate can be used with high confidence to perform thousands of iterations in seconds.

Conversely, for topological redesigns---such as a complete overhaul of the underbody or a change in the vehicle's greenhouse shape---the surrogate is often operating near or beyond its training distribution. Here, the risk of a ``blind spot'' is high, and the surrogate should be used only for initial screening, with mandatory CFD validation for any promising candidates. We identify three primary regimes for surrogate deployment:
\begin{itemize}
    \item \textbf{Exploratory Screening:} High-throughput sweeps to identify trends and eliminate obviously poor designs.
    \item \textbf{Local Optimization:} Fine-tuning of parametric features within a known design envelope.
    \item \textbf{Uncertainty-Aware Filtering:} Using the surrogate to decide which designs are ``safe'' to skip and which require high-fidelity verification.
\end{itemize}

Final sign-off for regulatory submission or safety-critical performance targets almost always requires high-fidelity CFD or wind tunnel testing. The surrogate's role is to ensure that the designs reaching that final stage are already highly optimized, rather than using the expensive final stage for basic discovery.

\subsection{Reliability and Uncertainty-Aware Escalation}
Accordingly, we treat ranking reliability, stability, and auditability as first-class engineering objectives---not afterthoughts. The goal is not to replace CFD entirely but to provide a high-throughput filter that can be trusted to navigate the design space. This requires explicit contracts for data provenance and preprocessing to ensure the model is not being asked to extrapolate into regimes where its training data is invalid \cite{rezaeiravesh2021uqit,smrs2025}.

Crucially, the system must support uncertainty-aware escalation \cite{chu2023uqcfd,uq4cfd2024,practicalguideuq2025}. When the model encounters a geometry that is significantly out-of-distribution or when its internal confidence intervals exceed a predefined threshold, it must have the ``right to remain silent'' and hand the decision back to the high-fidelity CFD pipeline. This hybrid approach ensures that the speed of ML is leveraged where possible, while the rigor of physics-based simulation is maintained where necessary. For heavy-duty vehicles, where drag reduction translates directly to significant fuel savings, the cost of a missed optimization opportunity is high, making robust uncertainty quantification a prerequisite for industrial adoption \cite{liu2024truck}.

\section{Famou-Agent and Evolutionary Coding Agent}
\label{sec:fmagent}

\subsection{Notation and Problem Statement}
We formalize the industrial surrogate discovery task as a constrained optimization problem over the space of executable programs. Let $\mathcal{D} = \{(\mathbf{x}_i, y_i)\}_{i=1}^{N}$ be a dataset where $\mathbf{x}_i \in \mathcal{X}$ represents a geometric descriptor (e.g., a 3D mesh or point cloud of a vehicle) and $y_i = C_{d,i} \in \mathbb{R}$ is the corresponding drag coefficient obtained from high-fidelity CFD solvers. The goal is to discover a surrogate function $f_\theta: \mathcal{X} \rightarrow \mathbb{R}$ that maximizes a ranking metric, such as the Spearman correlation coefficient $\rho$, while adhering to strict engineering contracts.

In industrial design, absolute error is often secondary to ranking stability; engineers require the surrogate to correctly order design candidates. Thus, we define the ranking loss $\mathcal{L}_{\mathrm{rank}}(f_\theta, \mathcal{D})$ based on pairwise ordering. Furthermore, we introduce an abstention rule: the surrogate should provide a prediction only when its internal uncertainty $\sigma(\mathbf{x})$ is below a predefined threshold $\sigma_{\max}$. The formal problem is stated as:
\begin{equation}
    \min_{f_\theta \in \mathcal{P}} \mathcal{L}_{\mathrm{rank}}(f_\theta, \mathcal{D}) \quad \mathrm{subject} \,\, \mathrm{to} \quad \mathcal{C}(f_\theta) = 1
\end{equation}
where $\mathcal{P}$ is the space of executable training programs and $\mathcal{C}(\cdot)$ is a boolean contract function that evaluates (i) data leakage, (ii) resource consumption, and (iii) reproducibility.

\subsection{Famou-Agent: The Foundation for Industrial Coding Agents}
This work builds on \textbf{Famou-Agent}, a self-evolving agent framework that iteratively proposes, executes, and revises code under automated evaluation \cite{li2025fmagent}. In an industrial setting, an agent framework is more than a prompting template: it standardizes (i) task decomposition, (ii) code/tool execution, (iii) evaluation contracts, and (iv) memory of failures and successes. Famou-Agent is a good fit for engineering pipelines because it treats the workflow as an executable system with explicit interfaces, not as a one-shot prompt. The agent writes code, runs it, observes failure modes (data issues, training crashes, evaluation leakage), and improves under a fixed harness. This echoes the general trajectory of tool-using and coding agents \cite{yao2022react,schick2023toolformer,yang2024sweagent}, but differs in emphasis: our output must be a reproducible pipeline with artifacts that can be re-run and audited.

From a systems perspective, Famou-Agent enables distributed asynchronous execution and evaluator-driven feedback loops \cite{li2025fmagent}. We adopt the same philosophy for surrogate discovery: a candidate is not a model checkpoint, but a versioned training program with (a) deterministic preprocessing, (b) split/leakage checks, (c) multi-seed evaluation, and (d) provenance metadata that makes results traceable \cite{atlasml2025,reprodl2025,smrs2025}.

\subsection{Evolutionary Objects and Operations}
To move beyond simple hyperparameter optimization \cite{feurer2020autosklearn2,hutter2024automl,shen2024automl}, we define a formal evolutionary framework where the agent operates on complex coding artifacts.

\subsubsection{Genome and Phenotype Representation}
The \textbf{Genome} $\mathcal{G}$ of our industrial coding agent is a composite object consisting of:
\begin{itemize}
    \item \textbf{Pipeline Configuration:} A structured specification (YAML/JSON) defining the DAG of operations, including data sources, solver versions, and resource constraints.
    \item \textbf{Python Code Artifacts:} A collection of modular scripts implementing custom data transformations, model architectures, loss functions, and evaluation metrics.
\end{itemize}
The \textbf{Phenotype} $\mathcal{P}$ is the \emph{executable training and evaluation program} instantiated from the genome. Unlike traditional EA where the phenotype is a static vector or fixed graph, our phenotype is a dynamic process that interacts with the industrial environment (e.g., CFD solvers, HPC schedulers).

\subsubsection{Mutation and Recombination Taxonomy}
The agent employs a specialized taxonomy of variation operators designed for ML pipeline engineering:
\begin{itemize}
    \item \textbf{Data Pipeline Edits:} Mutations that inject geometry canonicalization, outlier handling, or solver-version drift compensation.
    \item \textbf{Model Swaps:} Recombination of architectural blocks (e.g., replacing a standard MLP with a GNN or Point-Cloud backbone \cite{singh2026car}).
    \item \textbf{Loss Function Evolution:} Modification of the objective function, such as adding physics-informed constraints or uncertainty quantification (UQ) heads.
    \item \textbf{Split Policy Guards:} Evolution of the data splitting logic to prevent temporal or spatial leakage, ensuring ranking-stable decisions.
\end{itemize}

\subsubsection{Operator Semantics and Risk-Aware Selection}
To keep mutation behavior interpretable and auditable, we encode each operator as a
textual policy that links a code action to its expected benefit, dominant failure mode,
and corresponding contract gate. In practice, \textbf{data pipeline edits} (e.g., mesh
normalization and canonicalization adjustments) are used to improve convergence and
cross-program consistency, but they are monitored by determinism and geometry-integrity
checks to avoid silent distortion. \textbf{Model swaps} (e.g., replacing an MLP with a
GNN or point-cloud backbone) are used to inject stronger geometric inductive bias, while
resource gates guard against memory overflow and latency regressions on large meshes.
\textbf{Loss evolution} operators (e.g., adding physics penalties or ranking-aware terms)
are used to improve extrapolation and decision consistency, but they are coupled to
multi-seed variance checks to detect unstable training dynamics. Finally, \textbf{split
policy guards} enforce temporal/topological separation to prevent leakage; these operators
may reduce effective training volume, but leakage detectors and holdout discipline ensure
that ranking performance remains decision-valid rather than artifact-driven.

This textual operator specification replaces static lookup tables with executable intent:
every mutation proposal must state what it is trying to improve, which failure it may
introduce, and which gate will reject it if that failure appears. As a result, exploration
remains diverse without becoming unconstrained, and the search process is easier to audit
in industrial reviews.

\subsection{The Evolution Loop}
The core of the agent's operation is a population-based improvement loop, formalized in Algorithm~\ref{alg:evolution_loop}.

\begin{algorithm}[ht]
\caption{Industrial Evolution Loop}
\label{alg:evolution_loop}
\small % 如果太挤可以缩小一点点
\begin{list}{}{\leftmargin=2em \labelwidth=1.5em \labelsep=0.5em \itemsep=0pt \parsep=0pt \topsep=0pt}
    \item[\textbf{1:}] \textbf{Initialize:} Generate initial population $\mathcal{P}_0$ using diverse modeling theses (e.g., rendering-based \cite{song2023surrogate}).
    \item[\textbf{2:}] \textbf{while} budget remains \textbf{do}
    \item[\textbf{3:}] \quad \textbf{Evaluate:} For each candidate $c \in \mathcal{P}_t$:
    \item[\textbf{4:}] \quad\quad Run evaluation harness (folds, seeds, determinism checks).
    \item[\textbf{5:}] \quad\quad Check hard contract gates (leakage, resource limits).
    \item[\textbf{6:}] \quad\quad Compute multi-objective fitness $F(c)$.
    \item[\textbf{7:}] \quad \textbf{Select:} Identify elites and candidates for culling based on $F(c)$.
    \item[\textbf{8:}] \quad \textbf{Trace:} Record failure modes into \textit{Constraint Memory} to guide future mutations.
    \item[\textbf{9:}] \quad \textbf{Variate:} Apply mutation/recombination to elites to produce $\mathcal{P}_{t+1}$.
    \item[\textbf{10:}] \quad \textbf{Migrate:} Exchange elites between islands (if using Island Model).
    \item[\textbf{11:}] \textbf{end while}
    \item[\textbf{12:}] \textbf{return} Best reproducible pipeline and its provenance metadata.
\end{list}
\end{algorithm}

\subsection{Convergence and Quality-Diversity}
To ensure robust exploration of the modeling space, we incorporate principles from Quality-Diversity (QD) algorithms, specifically MAP-Elites \cite{mouret2015mapelites}. Instead of searching for a single global optimum, the agent maintains an archive of diverse, high-performing pipelines categorized by behavioral descriptors (e.g., model complexity vs. inference latency). The QD-Score, defined as the sum of fitness values across all cells in the archive, serves as a global measure of the agent's discovery progress.

Our approach also draws inspiration from FunSearch \cite{romeraparedes2024funsearch}, utilizing skeleton-based code templates where the LLM-based agent performs mutations on specific functional blocks. For multi-objective selection, we employ a strategy similar to NSGA-II \cite{deb2002nsgaii}, using non-dominated sorting to identify the Pareto front of pipelines that balance accuracy, reliability, and complexity. While formal convergence proofs for such high-dimensional code spaces are non-trivial, we hypothesize that under mild assumptions regarding the mutation operator's reachability and the evaluator's consistency, the archive quality is monotonically non-decreasing over generations.

\subsection{Selection and Fitness in Industrial Context}
Selection is multi-objective and reliability-aware. The fitness function $F(c)$ is defined as:
\begin{equation}
    F(c) = \omega_1 \cdot \mathrm{Accuracy} + \omega_2 \cdot \mathrm{Reliability} - \omega_3 \cdot \mathrm{Complexity} - \mathrm{Penalty}(\mathrm{Contract\_Violations})
\end{equation}
where \textbf{Reliability} accounts for variance across seeds and folds, and \textbf{Contract Violations} act as hard gates (auto-reject). If a candidate uses forbidden features or leaks holdout information, it is discarded regardless of apparent accuracy. This framing aligns the search with what engineers actually need: predictable pipelines and trustworthy comparisons rather than fragile leaderboard wins.

The individual terms are defined as follows:
\begin{itemize}
    \item \textbf{Accuracy:} Measured by the Spearman correlation $\rho$ on the validation set.
    \item \textbf{Reliability:} Defined as $1 - \sigma_{\mathrm{seeds}}$, where $\sigma_{\mathrm{seeds}}$ is the standard deviation of performance across multiple random seeds.
    \item \textbf{Complexity:} A penalty term proportional to the number of parameters or FLOPs, encouraging parsimonious models.
    \item \textbf{Penalty:} An indicator function where $\mathrm{Penalty}(v) = \infty$ if any hard contract gate is violated, and 0 otherwise.
\end{itemize}
The weights $\omega_i$ can be tuned via a small set of pilot experiments to reflect specific project priorities (e.g., prioritizing reliability for safety-critical components).

\subsection{Worked Example: One Generation}
To illustrate the agent's operation, consider a hypothetical generation starting with three candidate pipelines:
\begin{itemize}
    \item \textbf{Candidate A (Baseline):} A standard MLP operating on global geometric features.
    \item \textbf{Candidate B (PointNet):} A point-cloud based architecture \cite{singh2026car}.
    \item \textbf{Candidate C (GNN):} A graph neural network operating on the vehicle mesh.
\end{itemize}

Suppose the agent selects Candidate C for mutation. It analyzes the code and proposes adding an edge-attention mechanism to better capture local aerodynamic interactions. In the evaluation phase, the harness runs Candidate C' (the mutated version) across five folds and three seeds. During this process, the \emph{Leakage Detector} contract gate identifies that a specific geometric feature in the new attention block inadvertently includes information from the CFD solver's internal state, violating the temporal split policy. Consequently, Candidate C' is assigned a fitness of $-\infty$ and discarded.

Simultaneously, Candidate B shows high accuracy but also high variance across seeds. The agent records this failure mode into the \emph{Constraint Memory}. In the selection step, Candidate A remains the elite due to its high reliability, while Candidate B is flagged for "stability-focused mutation" in the next generation. Finally, the best version of Candidate A is migrated to a neighboring island to seed a new modeling thesis.

\subsection{Evaluation Harness and Traceability}
The evaluation harness is designed for industrial rigor:
\begin{itemize}
    \item \textbf{Folds and Seeds:} Mandatory multi-fold cross-validation with multiple random seeds to ensure statistical significance.
    \item \textbf{Determinism Checks:} Automated verification that the pipeline produces identical results given the same seed and environment.
    \item \textbf{Failure-Rate Accounting:} Tracking the frequency of transient failures (e.g., OOM, solver timeouts) as a first-class metric.
    \item \textbf{Memory/Traceability:} Every failure is analyzed by the agent to extract constraints (e.g., ``do not use feature X with solver version Y''), which are stored in a persistent memory to prune the search space \cite{li2025fmagent}.
\end{itemize}

\paragraph{Lifecycle Tightening, Diversity, and Ecosystem Context.} In practice, the harness is used with progressive contract tightening: early generations prioritize broad exploration under relatively lenient thresholds, while later generations enforce stricter latency, stability, and compliance limits. To avoid premature convergence, we run an island-style search with periodic elite migration and small random immigrant rates so different modeling theses can cross-pollinate without collapsing diversity. This design is consistent with broader SciML agentic tooling trends \cite{foameagent2025,scimlagents2025,agenticsciml2025,codexscientist2025}, and aligns with public evidence from Famou-Agent benchmark results \cite{li2025fmagent}.

\section{Application: Evolving a Drag-Prediction Surrogate} \label{sec:application}

\subsection{Problem Setting}
With the evolutionary machinery defined, we ground it in a concrete industrial task:
evolving a surrogate $f(\mathbf{x})\rightarrow\hat{C}_{\mathrm{d}}$ for
comparison-heavy drag optimization. In production programs, drag remains a first-order
driver of efficiency and electric-vehicle range, so candidate ranking quality is often more
important than small gains in pointwise error \cite{taira2024ev,sae2024ml}. The surrogate
is therefore treated as a governed product component, not a standalone benchmark model: it
must preserve ordering among close designs, remain stable across seeds and splits, and
respect strict auditability and safety constraints.

The deployment objective is to reduce expensive CFD dependence while preserving
decision-grade reliability. We use heterogeneous public and industrial data and related high-fidelity aerodynamic
benchmarks.
This multi-source setting forces the agent to discover modeling motifs that transfer across
vehicle topologies and operating regimes, while the evaluation contract ensures that any
throughput gains do not come from leakage or fragile shortcuts \cite{witherden2024pyfr,rumsey2025,pfaff2021meshgraphnets}.

\subsection{Data and Labels: The Governance Contract}
Surrogate quality is bounded by data quality, provenance, and leakage control. We treat the
dataset as a versioned asset under an explicit governance contract so every sample used in
evolution is traceable, validated, and compliant with program safety standards. This
contract functions as the agent's operational source of truth and prevents fragile gains
from low-quality or inconsistent data.

\paragraph{Data Provenance and Environment Specification.} Each dataset is paired with a
compact Dataset Card \cite{gebru2021datasheets} that records geometry source, solver
version/settings used for label generation, and known distribution limits. We also retain
core flow-condition metadata (e.g., Reynolds number and turbulence setup) to align training
and deployment regimes.

\paragraph{Label Fidelity and Geometry Integrity.} Ground-truth labels ($C_{\mathrm{d}}$)
are generated with high-fidelity CFD (typically RANS or DES) under fixed convergence and
mesh-independence criteria. In parallel, geometry canonicalization enforces consistent
coordinates, mesh density policy, and feature-preserving preprocessing so the model learns
physics-relevant structure rather than meshing artifacts.

\paragraph{Leakage Policy (Hard Prohibitions).} We ban any feature derived from solver
internal states, project/model identifiers, metadata that can encode labels, or information
from holdout/certification splits. These constraints are enforced by automated checks in
the evaluation harness, and violations trigger immediate rejection.

\paragraph{Split Policy and Generalization.} Random splits are disallowed because they can
inflate performance under temporal/spatial correlation. We enforce hard splits by vehicle
family, design phase, or topology so evaluation reflects forward-prediction reality in
industrial programs. This keeps ranking performance tied to true generalization rather than
dataset overlap artifacts.

\subsection{Integrated Multi-Fidelity Training and Contracted Evaluation}
In practice, we couple multi-fidelity data integration with executable evaluation gates in
a single loop. Training data combines abundant low-cost RANS labels with sparse
high-fidelity LES and wind-tunnel samples; candidate pipelines learn global trends from
low-fidelity sources while high-fidelity samples anchor local accuracy in difficult
separated-flow regimes \cite{multifidelity2025,pisaroni2017mlmc,chu2023uqcfd}.

Every candidate is judged by a contract-as-code harness rather than headline accuracy
alone. A pipeline must pass reproducibility, resource, leakage, and stability gates before
selection: artifacts and environments must be replayable, compute budgets respected,
forbidden features absent, and cross-seed variance controlled. Uncertainty is mandatory;
when confidence is low, the system escalates to high-fidelity CFD instead of committing a
low-trust prediction.

The contract is lifecycle-adaptive. Early generations use looser thresholds to encourage
exploration; refinement stages tighten latency and robustness constraints; certification
stages require blind-test performance and audit-grade compliance for deployment
\cite{sae2024ml,smrs2025}. This unified design improves simulation-budget efficiency
while preserving industrial reliability and governance traceability.

\section{Performance Evaluation and Attribution}
\label{sec:analysis}
\subsection{Overall Multi-Model Performance}
\label{sec:overall_performance}

To evaluate the robustness and generalizability of the proposed evolutionary framework, we conducted a comprehensive benchmark across eight distinct LLM-driven evolutionary operators. Each operator was tasked with evolving a high-fidelity surrogate pipeline for $C_d$ prediction under identical iteration counts and compute budgets. To quantify the "evolutionary fitness" of each generated program, we define a \textbf{Combined Score} ($S$) as the primary evaluation metric. This score is designed to prioritize the physical consistency of aerodynamic trends while maintaining numerical accuracy, calculated as follows:

\begin{equation}
    S = \alpha \cdot \text{Acc}_{\text{sign}} + \beta \cdot \frac{1}{1 + \text{RMSE}} + \gamma \cdot \frac{1}{1 + \text{MAE}}
    \label{eq:combined_score}
\end{equation}

where $\text{Acc}_{\text{sign}}$ denotes the Sign Accuracy of the predicted pressure-drag changes ($\Delta C_d$) between design pairs. This weighting scheme explicitly favors models that correctly capture the direction of aerodynamic variation, which is of higher engineering utility in the iterative vehicle design process than mere numerical fitting. Table~\ref{tab:multi_model_results} summarizes the empirical results, reporting the mean performance metrics across three independent evolutionary seeds for each operator.

\begin{table}[htbp]
\caption{Mean Predictive Performance Metrics Across Anonymous Evolutionary Operators (n=3).}
\centering
\begin{tabular}{lcccc}
\toprule
Evolutionary Operator & Combined Score $\uparrow$ & Sign Accuracy $\uparrow$ & MAE $\downarrow$ & RMSE $\downarrow$ \\
\midrule
gemini-3.0-pro & \textbf{0.9335} & \textbf{0.9180} & 0.0279 & 0.0354 \\
gpt-5.2 & 0.8880 & 0.8671 & \textbf{0.0261} & \textbf{0.0322} \\
gemini-2.5-pro & 0.8832 & 0.8617 & 0.0283 & 0.0352 \\
qwen3-coder-480b & 0.8702 & 0.8454 & 0.0285 & 0.0353 \\
deepseek-r1 & 0.8437 & 0.8121 & 0.0276 & 0.0339 \\
deepseek-v3.2-think & 0.8241 & 0.7917 & 0.0447 & 0.0546 \\
deepseek-v3.2 & 0.6269 & 0.5471 & 0.0519 & 0.0637 \\
\bottomrule
\end{tabular}
\label{tab:multi_model_results}
\end{table}

\begin{figure}[t]
    \centering
    \includegraphics[width=0.8\textwidth]{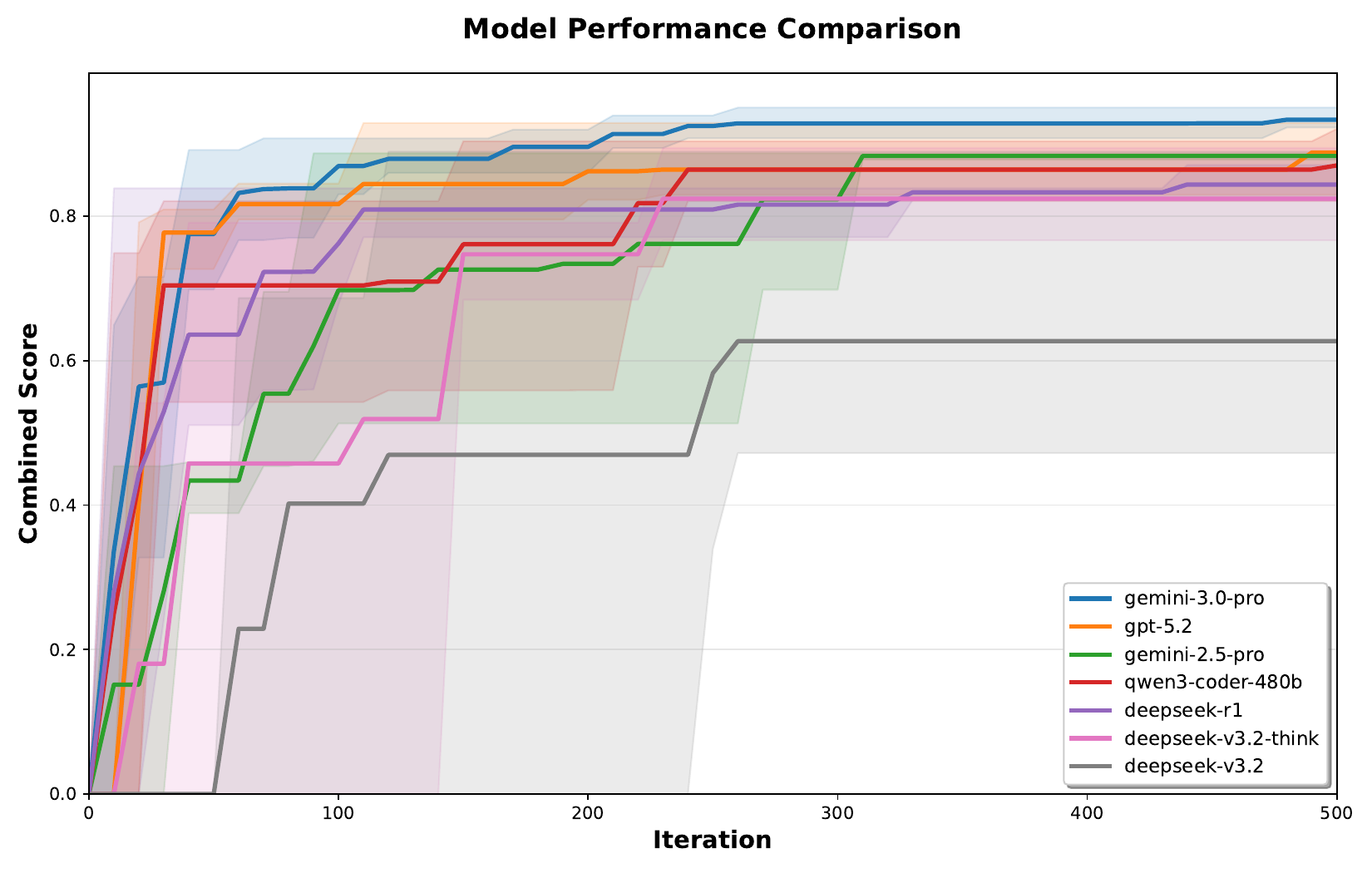}
    \caption{Comparative evolutionary trajectories showing the mean combined score across three independent runs, with shaded areas representing the range between the maximum and minimum performance observed at each iteration.}
    \label{fig:evolution_trajectory}
\end{figure}

\textbf{Statistical Observations:} The experimental results in Table~\ref{tab:multi_model_results} demonstrate significant variability in "evolutionary fitness" across different backends. Gemini-3.0-pro achieves the highest Combined Score (0.9335), primarily driven by its superior Sign Accuracy (91.80\%), which reflects its robust capability in capturing correct physical trends. While gpt-5.2 achieves the lowest absolute error (MAE: 0.0261; RMSE: 0.0322), its slightly lower Combined Score is due to the weighted scoring mechanism that prioritizes Sign Accuracy—a critical metric for ensuring the physical consistency of aerodynamic surrogates over mere numerical fitting.

As further illustrated by the evolutionary dynamics in Figure~\ref{fig:evolution_trajectory}, four distinct convergence patterns emerge:
\begin{itemize}
    \item \textbf{Frontier Reasoning Dominance:} Gemini-3.0-pro demonstrates an elite "fast-start" capability, effectively breaking through initial search plateaus within the first 100 iterations. It reaches a performance level in its early stage that most other operators fail to attain even at the end of the 500-iteration cycle.
    
    \item \textbf{Delayed High-Fidelity Emergence:} Initial progress is not always predictive of final efficacy. Gemini-2.5-pro displays significant "cold-start" latency with a measured early trajectory, yet it maintains a consistent positive gradient throughout the search, eventually securing a top-tier rank (3rd overall) by prioritizing steady refinement over radical mutation.
    
    \item \textbf{Baseline Performance Gap:} Deepseek-v3.2, representing the most basic model architecture among the operators, displays a stark performance gap. It struggled to identify any valid solutions until approximately the 50th iteration, and its final score remains significantly lower than that of the other frontier models.
\end{itemize}

In summary, these findings confirm that navigating the high-dimensional aerodynamic search space requires a synergistic combination of robust code generation for executable implementation and deep logical reasoning for meaningful architectural evolution.

\subsection{Evolutionary Attribution and Trajectory Analysis}
\label{sec:evolutionary_attribution}

To further elucidate how the Famou Agent facilitates performance gains, we perform a deep-dive attribution analysis on a representative evolutionary run. As illustrated in Figure~\ref{fig:lineage_plot}, the evolutionary trajectory of the best-performing program is visualized by marking key mutation events. Each annotation $(s, i, g)$ on the plot represents the \textbf{Combined Score} ($s$), the \textbf{Iteration} count ($i$), and the \textbf{Generation} count ($g$), respectively. The evolution process can be categorized into four distinct strategic epochs based on this lineage:

\begin{figure}[htbp]
    \centering
    \includegraphics[width=0.8\textwidth]{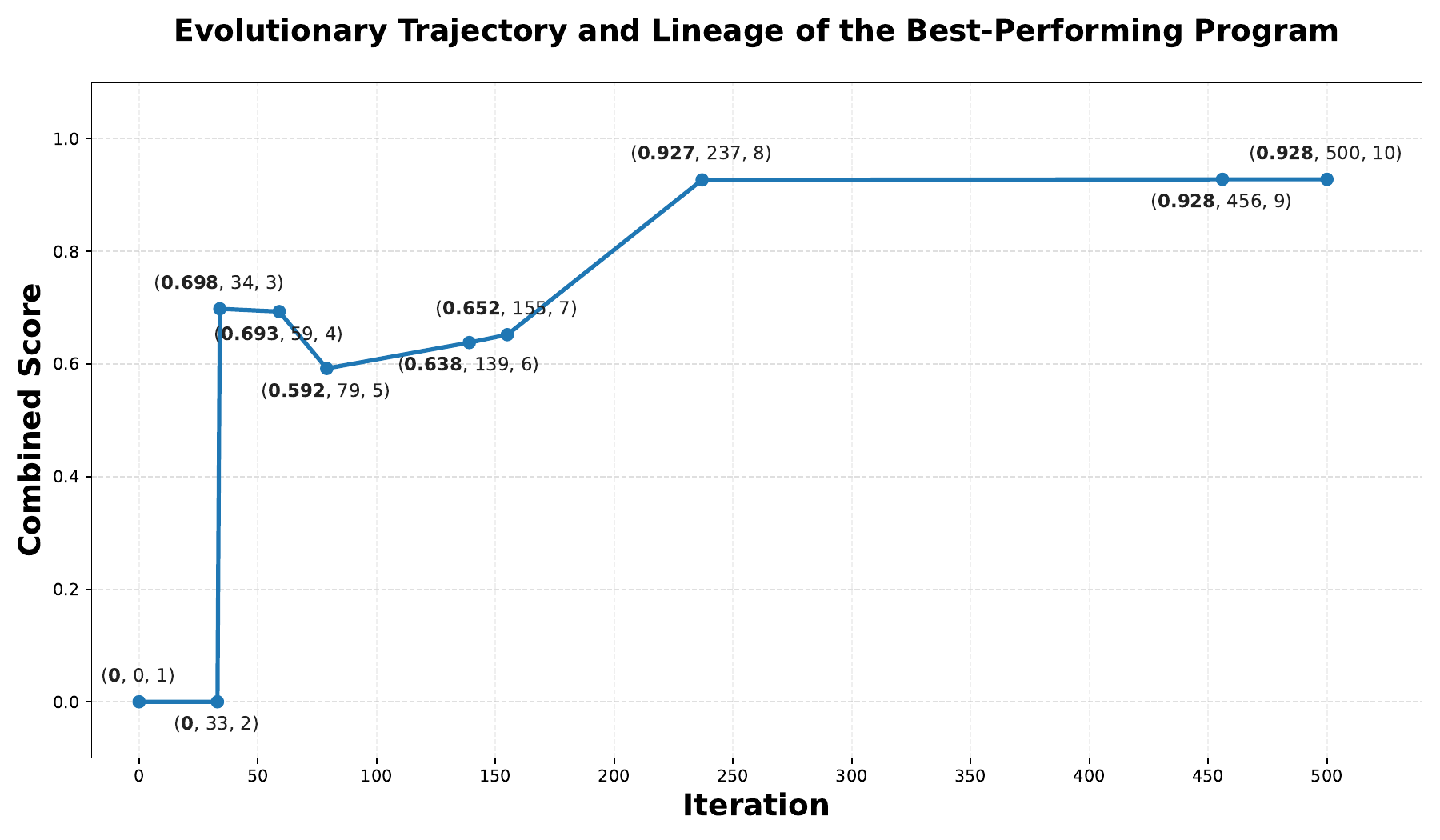}
    \caption{Evolutionary Trajectory and Lineage of the Best-Performing Program. The plot explicitly labels all data points along the evolutionary path, where each tuple $(s, i, g)$ represents the Combined Score, Iteration, and Generation.}
    \label{fig:lineage_plot}
\end{figure}

\textbf{1. Initial Phase (Score 0):} The initial phase was characterized by a score of zero due to execution-level failures. Early mutations suffered from structural flaws, such as the \texttt{NameError} for global constants and parameter naming mismatches in \texttt{TransformerConv}. The agent's breakthrough during this stage was not in physics, but in \textit{robustness}: by fixing naming conventions and recovering utility constants, it transitioned the pipeline from "runtime failure" to "executable training," enabling edge features to participate in backpropagation for the first time.

\textbf{2. Feasibility Breakthrough (Score 0 $\rightarrow$ 0.698):} A critical inflection point occurred at iteration 34, represented by the jump to $0.698$. The evolutionary process pivoted upon the realization that absolute $C_d$ regression is inherently difficult due to the high sensitivity of 3D geometry, subsequently refactoring the objective function from a pure MSE-based regression to a multi-task framework incorporating a Pairwise Ranking Loss. By penalizing sign inconsistencies between design pairs, the model successfully captured relative aerodynamic trends, resulting in the first dramatic fitness leap. To support this under limited VRAM, the agent simultaneously implemented Mixed Precision Training.

\textbf{3. Stagnation and Exploration Phase (Score 0.698 $\rightarrow$ 0.652):} Between iterations 34 and 155, the fitness score exhibited sustained oscillations and stagnation around the 0.6 baseline. Although the agent experimented with various graph convolution kernels (e.g., custom \texttt{GeoEdgeLayers}) and introduced features like node centering, the pipeline failed to achieve a significant performance breakthrough. This stage represents a typical local optimum, where the search process was hindered by the inherent trade-off between model depth and gradient stability.

\textbf{4. Global Optimization Breakthrough (Score 0.652 $\rightarrow$ 0.927):} The most significant performance jump occurred at iteration 237, reaching $0.927$. This breakthrough was triggered by an \textbf{Island Migration}, which effectively rescued the search from local optima. The migrated agent introduced a GeoTransformerConv architecture, integrating relative geometric information directly into the multi-head attention mechanism. Coupled with a transition to a LogSigmoid-based ranking loss with adaptive thresholds, the pipeline achieved a decisive breakthrough, eventually stabilizing at $0.928$.

In conclusion, the trajectory analysis reveals that the Famou Agent performs more than simple hyperparameter tuning by iteratively solving engineering bottlenecks ranging from memory management to loss function topology, while also leveraging population-level diversity to overcome stagnation in the high-dimensional design space.

\subsection{Ablation Study of Evolved Strategies}
\label{sec:ablation_study}

To quantify the individual contributions of the core architectural components within the Famou Agent, we conduct a systematic ablation study. We compare the "Full Method" against three constrained variants: (1) \textbf{w/o Feedback}, which removes LLM-driven context such as error reflections; (2) \textbf{w/o Island Model}, which disables the population-based island migration mechanism; and (3) \textbf{w/o Adaptive Sampling}, which replaces the adaptive sampling strategy with a conventional Top-K greedy sampling approach (pure exploitation). Deepseek-r1 was selected as the representative base operator for the ablation study to provide a balanced performance baseline with sufficient headroom for observing both potential functional degradation and synergistic enhancements.

\begin{figure}[htbp]
    \centering
    \includegraphics[width=0.8\textwidth]{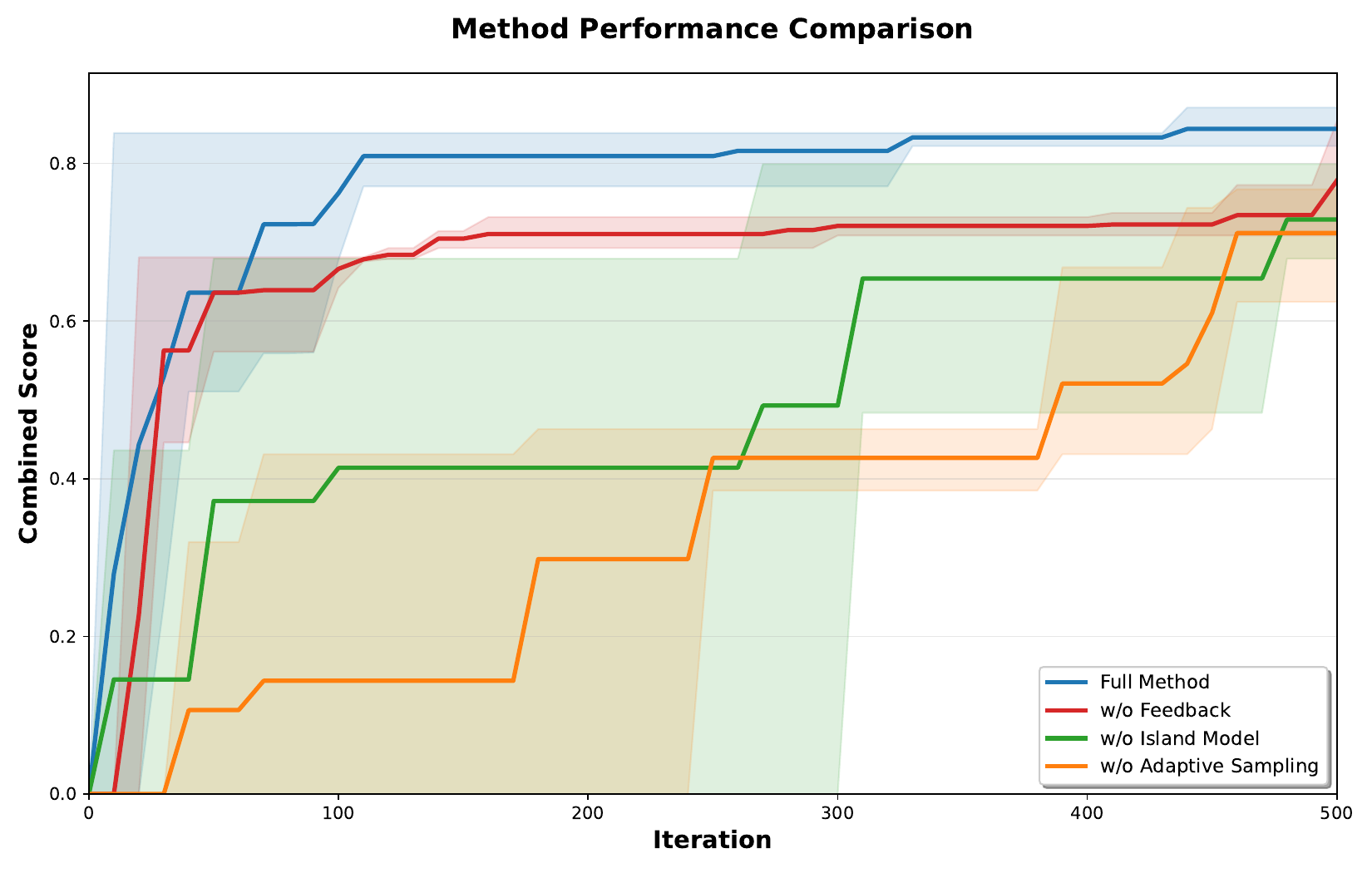}
    \caption{Ablation analysis of the Famou Agent framework using the intermediate-level model (Deepseek-r1). The trajectory plots illustrate the mean Combined Score averaged over three independent runs, with the shaded regions denoting the min-max performance bounds across these trials.}
    \label{fig:ablation_comparison}
\end{figure}

The comparative results, as visualized in Figure~\ref{fig:ablation_comparison}, reveal that the Full Method consistently outperforms all ablated variants in both convergence speed and final solution quality. A striking observation is that the \textbf{Full Method} (average score: 0.8437) achieves a performance level within the first 100 iterations that the other strategies fail to reach even at the end of their 500-iteration cycles. This "fast-start" capability underscores the synergistic effect of combining directed feedback with diverse population search.

The analysis of specific modules highlights their distinct roles in the evolutionary process:
\begin{itemize}
    \item \textbf{Impact of Adaptive Sampling:} Removing adaptive sampling (\textbf{w/o Adaptive Sampling}) leads to the most significant performance degradation, with the average score dropping to 0.7117. The trajectory exhibits an extremely sluggish startup process and discrete step-like increases. This suggests that in the complex non-convex search space of 3D aerodynamic models, a pure-exploitation Top-K strategy, due to the lack of an effective exploration mechanism, finds it extremely difficult to discover potential optimization directions. This results in a search efficiency far lower than other variants and an ultimate failure to converge to a high-level solution.
    
    \item \textbf{Impact of the Island Model:} The variant \textbf{w/o Island Model} achieves a final average score of 0.7287. Notably, the score stagnates between iterations 110 and 270, indicating that the population mechanism lacking island interactions is prone to getting trapped in local optima. Furthermore, this variant exhibits significant instability across multiple runs, with one independent experiment only yielding valid solutions after iteration 300. This further corroborates our finding in Section~\ref{sec:evolutionary_attribution} that interactions among islands are essential for triggering global optimization breakthroughs and maintaining population diversity.
    
    \item \textbf{Impact of Feedback:} The \textbf{w/o Feedback} variant (average score: 0.7782), which lacks contextual information such as LLM-driven error reflection, exhibits a significantly lower search efficiency. Due to the inability to learn from accumulated historical experience, the agent is prone to repeating previous errors. Compared to the Full Method, this results in a more stochastic mutation process that lacks clear directional guidance.
\end{itemize}

In summary, the ablation study demonstrates that while each component contributes to the overall success, the Adaptive Sampling strategy and Island Model strategy are the primary drivers for navigating high-dimensional engineering constraints. The integration of LLM-driven feedback further accelerates this process by providing a logical roadmap for architectural refinement.

\section{Engineering Impact and Deployment}
\label{sec:impact}

This section focuses on deployment: how to use the surrogate to accelerate engineering decisions without replacing high-fidelity Computational Fluid Dynamics (CFD). We recommend a ``screen-and-escalate'' operating model in which the surrogate handles high-throughput ranking, while CFD remains the judge of record for final validation. This division of roles preserves physics-grade confidence at decision time and directs limited high-fidelity budget toward the candidates that matter most.

\subsection{Operational Model and Daily Workflow}
In a practical vehicle program, engineers first generate large batches of parameterized geometry variants and evaluate them with the surrogate to obtain both predicted performance and calibrated uncertainty. Instead of sending every candidate to the HPC queue, the workflow escalates only low-confidence or out-of-envelope designs to CFD, together with a small top-ranked subset for confirmation. The resulting CFD labels are then fed back into the training set so the surrogate can be retrained or evolved against the latest solver settings and design language. In mature deployments, this closed loop can compress the design-to-shortlist cycle from weeks to hours while maintaining high ranking fidelity on contract-governed holdout evaluations. The central benefit is not merely speed in one run, but sustained throughput across the full optimization program.

\subsection{Governance, Integration, and Reliability}
Reliable adoption depends on governance and system discipline as much as predictive performance. A robust rollout typically begins in shadow mode, where surrogate outputs are logged but do not yet drive critical decisions, then progresses to partial automation with spot-checking, and finally to selective autonomous screening under explicit guardrails. At the infrastructure level, predictions must be fully traceable to model version, data version, preprocessing logic, and solver context through integration with PLM/PDM systems and auditable deployment pipelines \cite{atlasml2025,sae2024ml}. Operational reliability further requires continuous monitoring for data drift, concept drift, and solver-version drift, with automatic policy tightening when risk indicators rise \cite{smrs2025}. If validation failures accumulate, rollback and circuit-breaker mechanisms must disable the surrogate and return control to a CFD-only path until remediation is complete.

\subsection{Impact Measurement: KPIs and ROI}
To support deployment decisions with evidence, impact should be measured through a consistent KPI framework that links technical performance to engineering outcomes. Throughput gain quantifies how many more designs can be screened per hour relative to a CFD-only baseline, while escalation rate captures how much of the design stream still requires high-fidelity intervention under uncertainty controls. Ranking fidelity on top-$K$ candidates reflects the reliability of surrogate-guided prioritization, and resource-efficiency metrics translate this reliability into reduced HPC core-hours for reaching target performance. Time-to-decision captures the full wall-clock effect from batch generation to validated shortlist. We define return on investment as
\begin{equation}
    ROI = \frac{(N_{screened} \times Cost_{CFD}) - (Cost_{Surrogate} + Cost_{Validation})}{Cost_{Surrogate} + Cost_{Validation}}
\end{equation}
where $Cost_{Validation}$ includes escalation and top-$K$ confirmation runs. Beyond direct cost savings, teams should also report counterfactual impact by estimating which design improvements were discovered only because surrogate-based screening made broader exploration economically feasible.

\subsection{Safety Boundaries and Scope}
\label{sec:limits}
No surrogate is safe by default, and this is especially true under distribution shift caused by new geometries, updated solver stacks, or changing operating targets. For this reason, safety is treated as an operational property of the full system rather than an intrinsic property of the model. In practice, the deployment pipeline must continuously monitor for data drift, concept drift, and calibration degradation, and must enforce conservative escalation to CFD whenever uncertainty grows or out-of-distribution behavior is detected. This prevents plausible but unreliable predictions from silently contaminating optimization loops \cite{chu2023uqcfd,practicalguideuq2025,uq4cfd2024}.

The framework also defines explicit scope boundaries. It does not claim universal validity across all flow regimes, replacement of high-fidelity CFD for certification, immunity to adversarial geometric perturbations, or guaranteed convergence to globally optimal pipelines. These constraints are intentional and should be documented with versioned model cards, data sheets, and solver tags so that decisions remain auditable over time \cite{gebru2021datasheets,mitchell2019modelcards,li2025fmagent}. From a governance perspective, model lifecycle controls (including retraining/retirement policies), human-in-the-loop final sign-off, and clear fallback procedures are required for safe industrial use \cite{sae2024ml,reprodl2025}.

\section{Conclusion}

The results across trajectory analysis, ablation, and deployment evaluation indicate that this framework contributes an industrial process rather than only a better standalone model. In the representative evolutionary run, the agent progressed from execution-level failure to a high-performing regime (Combined Score near 0.93) by resolving pipeline robustness issues, shifting the objective toward ranking-aware learning, and exploiting island migration to escape local optima (Section~\ref{sec:evolutionary_attribution}). The ablation results further show that these gains are structural, not incidental: the Full Method (0.8437 average) outperforms all constrained variants, while removing adaptive sampling (0.7117), island migration (0.7287), or feedback context (0.7782) causes slower convergence and lower final quality (Section~\ref{sec:ablation_study}). Taken together, these findings support the claim that executable-contract evolution, diversity-preserving search, and feedback-guided mutation are all necessary to achieve stable, high-quality drag-surrogate discovery in high-dimensional automotive design spaces.

From an industrial enablement perspective, the framework is valuable because it turns drag prediction into a reliable decision-support layer for engineering throughput. The screen-and-escalate operating model allows teams to rank large design batches quickly, reserve CFD for top candidates and uncertain cases, and continuously improve the surrogate with newly validated labels, which in mature settings can compress design-to-shortlist cycles from weeks to hours while preserving ranking reliability (Section~\ref{sec:impact}). Equally important, deployment discipline---traceability, drift monitoring, rollback, and explicit scope boundaries---keeps this acceleration compatible with safety-critical development practice (Section~\ref{sec:limits}). In this sense, the core contribution is a practical blueprint for using self-evolving agents to increase design iteration capacity, improve CFD budget allocation, and maintain auditable confidence as vehicle programs and solver stacks evolve.

\section*{Acknowledgments}
The authors thank their institutional collaborators and compute support teams for enabling this work. Detailed funding and resource disclosures will be added in the camera-ready version.

%Bibliography
\bibliographystyle{unsrt}  
\bibliography{references}

\end{document}